\documentclass[10pt,twocolumn,letterpaper]{article}

\usepackage{iccv}
\usepackage{times}
\usepackage{epsfig}
\usepackage{graphicx}
\usepackage{grffile} % allow spaces and other special characters in graphic file names
\usepackage{amsmath}
\usepackage{amssymb}
\usepackage{subcaption} % use subcaption (don't load subfig too; it conflicts and can break \begin{document})
\usepackage{enumitem}
\setlist[itemize]{label=\textbullet}

\usepackage[pagebackref=true,breaklinks=true,letterpaper=true,colorlinks,bookmarks=false]{hyperref}

\iccvfinalcopy % Uncomment this line for the final submission

\begin{document}
\pagestyle{plain} % Enable page numbers

%%%%%%%%% TITLE
\title{A Probabilistic Framework for Improving Dense Object Detection in Underwater Image Data via Annealing-Based Data Augmentation}

\author{%
Eleanor Wiesler\textsuperscript{*,1,2}
\and
Trace Baxley\textsuperscript{*,1,2}
}

\maketitle

% Footnotes for author attributions (shown as *,1,2 in the author line)
\begingroup
\renewcommand{\thefootnote}{}%
\makeatletter\renewcommand{\@makefnmark}{}\makeatother%
\footnotetext{\textsuperscript{1} Harvard Department of Mathematics.}%
\footnotetext{\textsuperscript{2} Harvard Department of Computer Science.}%
\footnotetext{\textsuperscript{*} Joint first authors (equal contribution).}%
\endgroup

\thispagestyle{empty}

%%%%%%%%% ABSTRACT
\begin{abstract}
Object detection models typically perform well on images captured in controlled environments with stable lighting, water clarity, and viewpoint, but their performance degrades substantially in real-world underwater settings characterized by high variability and frequent occlusions.

In this work, we address these challenges by introducing a novel data augmentation framework designed to improve robustness in dense and unconstrained underwater scenes. Using the DeepFish dataset, which contains images of fish in natural environments, we first generate bounding box annotations from provided segmentation masks to construct a custom detection dataset. 

We then propose a pseudo–simulated annealing–based augmentation algorithm, inspired by the copy-paste strategy of Deng et al.~\cite{deng}, to synthesize realistic crowded fish scenarios. Our approach improves spatial diversity and object density during training, enabling better generalization to complex scenes. Experimental results show that our method significantly outperforms a baseline YOLOv10 model, particularly on a challenging test set of manually annotated images collected from live-stream footage in the Florida Keys. These results demonstrate the effectiveness of our augmentation strategy for improving detection performance in dense, real-world underwater environments.
\end{abstract}

%%%%%%%%% BODY TEXT
\section{Introduction}

Using fish detection algorithms is a significant challenge due to the most popular models being used for controlled research or for fish farms. In natural environments, water clarity, lighting conditions, and orientation inhibit these models from generalizing. 

There are some models in the literature that utilize YOLO for fish detection, but do not do so on natural environments but use controlled environments like tanks or fish farms instead \cite{cite-key}. Similarly, there are a small number of studies that focus on natural environments, but very few use CNN models. [site papers with natural environments but not cnn] Moreover, the models in the literature do not address the specific features of our test set that lead to decreased robustness (namely, interspersed crowding).

As a result, our aim is to address these limitations of these models by training a YOLO-based detection model using the DeepFish dataset \cite{deepfish}, which contains images of fish within their natural environment and with varying lighting conditions, generating bounding boxes from provided segmentation data. 

Through exploring the DeepFish dataset, there is an additional limitation to the segmentation data used for training -- there are too few fish in each image. Thus, we aim to address this discrepancy through fine-tuning via data augmentation. 
%-------------------------------------------------------------------------
\section{Related Work}

\subsection{Crowded Object Detection in Computer Vision}
There have been many studies on improving the performance of CNN models for object detection in crowded scenarios. For instance, the majority of models use Non-Maximum Supression (NMS) to deal with crowding in models as done in Luo et al \cite{Luo_2021}. However, this relies on simplifying the input data and is not accurate for counting purposes. These models also tend to suffer from a high false-positive rate. Other models that aim to improve robustness for crowded detection use a different loss function called focal loss to place less importance for foreground detections \cite{lin2018focallossdenseobject}. These models, however reveal that this loss function aids most in class disparity, and our problem does not involve classification. A recent adjustment to CNN models for crowding was introduced by Deng et al \cite{deng}, who introduced a copy-paste algorithm for data augmentation that led to an improvement all standard performance metrics for human detection. The algorithm is as follows: 

\begin{enumerate}
    \item \textbf{Select group centers.}
    \begin{itemize}
        \item Define a set \( C \) of group centers \( C = \{(x_1, y_1, s_1), \dots, (x_{|C|}, y_{|C|}, s_{|C|})\} \), where \( (x_i, y_i) \) are the coordinates and \( s_i \) is the normalized object size.
        \item The group number \( |C| \) is randomly sampled from the range \([0, N]\), where \( N \) is a predefined hyperparameter.
        \item Group centers are selected by sampling from the original objects in the image.
    \end{itemize}

    \item \textbf{Generate groups around each center.}
    \begin{itemize}
        \item For each group center \( c_i \in C \), generate a group \( \hat{G}_i \) of objects:
        \[
        \hat{G}_i = \{(x_1, y_1, s_1), \dots, (x_{|\hat{G}_i|}, y_{|\hat{G}_i|}, s_{|\hat{G}_i|})\}.
        \]
        \item The number of objects \( |\hat{G}_i| \) in each group is randomly sampled from the range \([0, M]\), where \( M \) is another hyperparameter.
        \item Enforce overlapping between each object \( g_{ij} \in \hat{G}_i \) and its corresponding group center \( c_i \).
    \end{itemize}

    \item \textbf{Simulate realistic crowdedness.}
    \begin{itemize}
        \item Object sizes in a group follow a Gaussian distribution:
        \[
        p(s_j | s_i, I) = \frac{1}{\sqrt{2\pi\sigma^2}} \exp\left( -\frac{(s_j - s_i)^2}{2\sigma^2} \right),
        \]
        where \( \sigma = 0.2 \) is a fixed standard deviation.
        \item Coordinates \( x \) and \( y \) are sampled from uniform distributions around the group center \( (x_i, y_i) \):
        \[
        x_j \sim U(x_i - \tau d_w, x_i + \tau d_w),
        \]
        \[
        y_j \sim U(y_i - \varepsilon d_h, y_i + \varepsilon d_h),
        \]
        where \( d_w \) and \( d_h \) are maximum allowable displacements, and \( \tau > 1 \) and \( \varepsilon > 1 \) control the degree of crowdedness.
    \end{itemize}
\end{enumerate}
We make significant adjustments to this algorithm for the development of our novel model for fish detection as will be discussed in 4.2.

\subsection{Simulated Annealing}
Simulated annealing is a probabilistic algorithm that optimizes by traversing from local minima with a small probability \cite{sim}. While the algorithm presented in future sections for our data augmentation does not actually use simulated annealing, many of its features are shared with simulated annealing algorithms (the small probability that a mask image will be further away from a center point). Thus, the algorithm is designed so that the model created from the CNN is robust to distinct patterns of fish clustering with our limited dataset. 

%-------------------------------------------------------------------------
\section{Dataset Preparation}
\subsection{DeepFish Dataset}

The DeepFish dataset \cite{deepfish} provided us with data for segmentation, localization (detection) and classification, and also provided a predetermined split for train, test, and validation data for each of the categories. 

Because the localization set had no predetermined annotations for data bounding boxes, for the sake of time we used the Segmentation data set because they included masks for the train and the validation data. Thus, we generated the bounding box using cv2's connected-components function, and generated .txt files based on the bounding box according to syntax of YOLOv10. A vizualization of this generation is below: 

\begin{figure}[htbp]
    \centering
    \includegraphics[scale = 0.6]{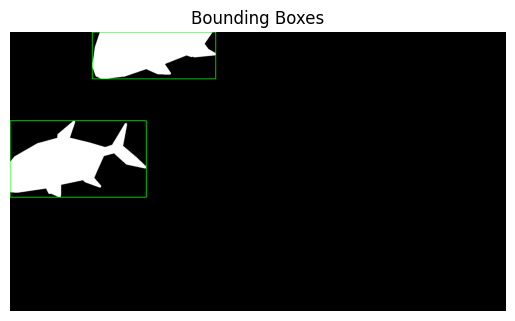} % Replace 'example-image' with your image file name
    \caption{Bounding Box generated from a validation set segmentation mask}
    \label{fig:your-label}
\end{figure}

\subsection{Data Augmented- DeepFish}

The original model was not robust to crowding in some of the images in our test set. Thus, we implemented the data augmentation algorithm relayed in Section 4. 

For the updated model, our base images were sourced from the original training set. Moreover, using the mask set, we were able to make a clean cut of the fish in our training set for random selection in a different folder for random segment generation. An example of one of the augmented images is shown below: 

\begin{figure}[htbp]
    \centering
    \includegraphics[scale = 0.12]{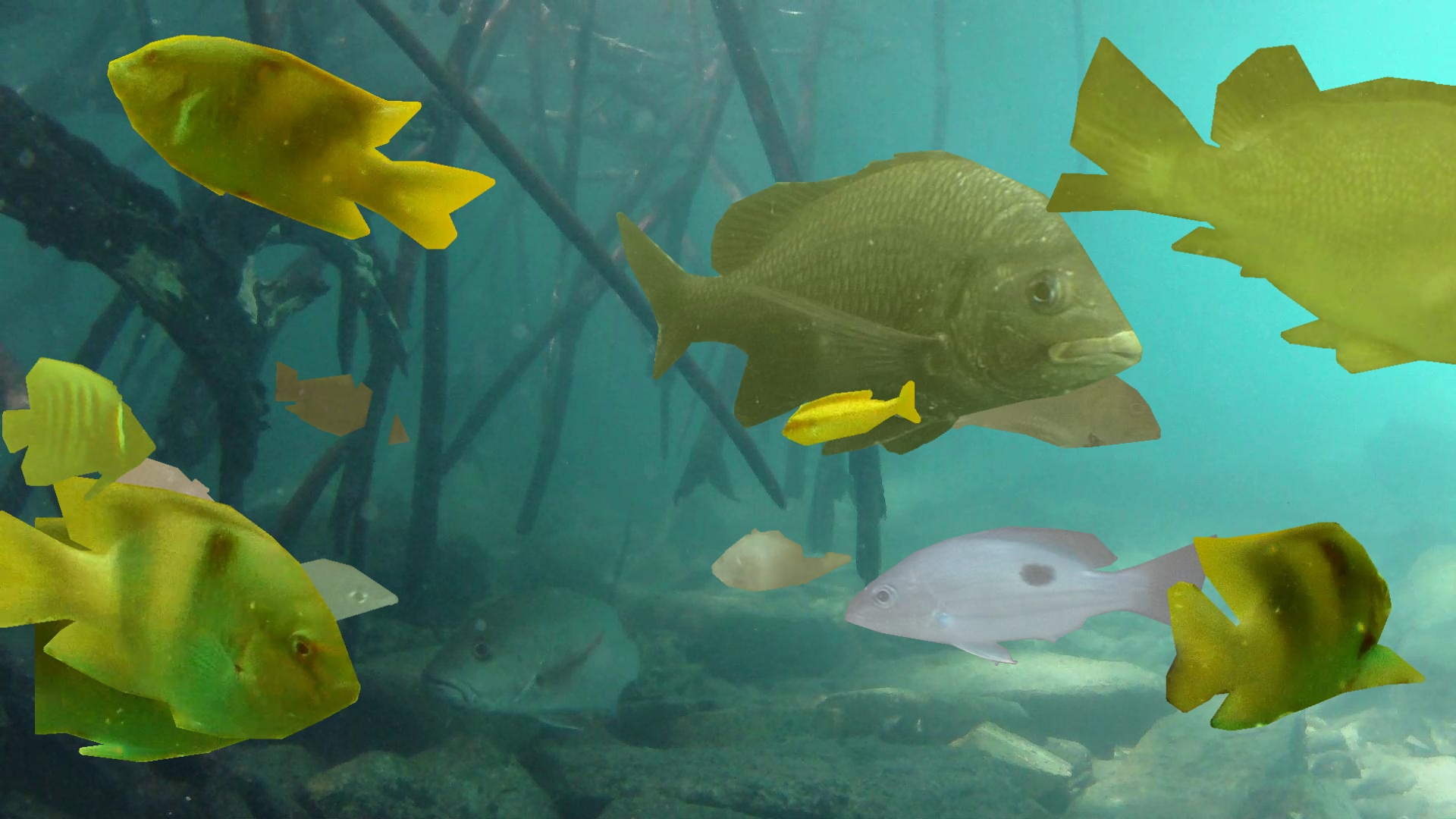} % Replace 'example-image' with your image file name
    \caption{Training image constructed using  adjusted Copy-Paste algorithm. Copy-pasted fish with augmented yellow-green color are a result of algorithm used for improved training.}
    \label{fig:your-label}
\end{figure}

\begin{figure}[htbp]
    \centering
    \includegraphics[scale = 0.1]{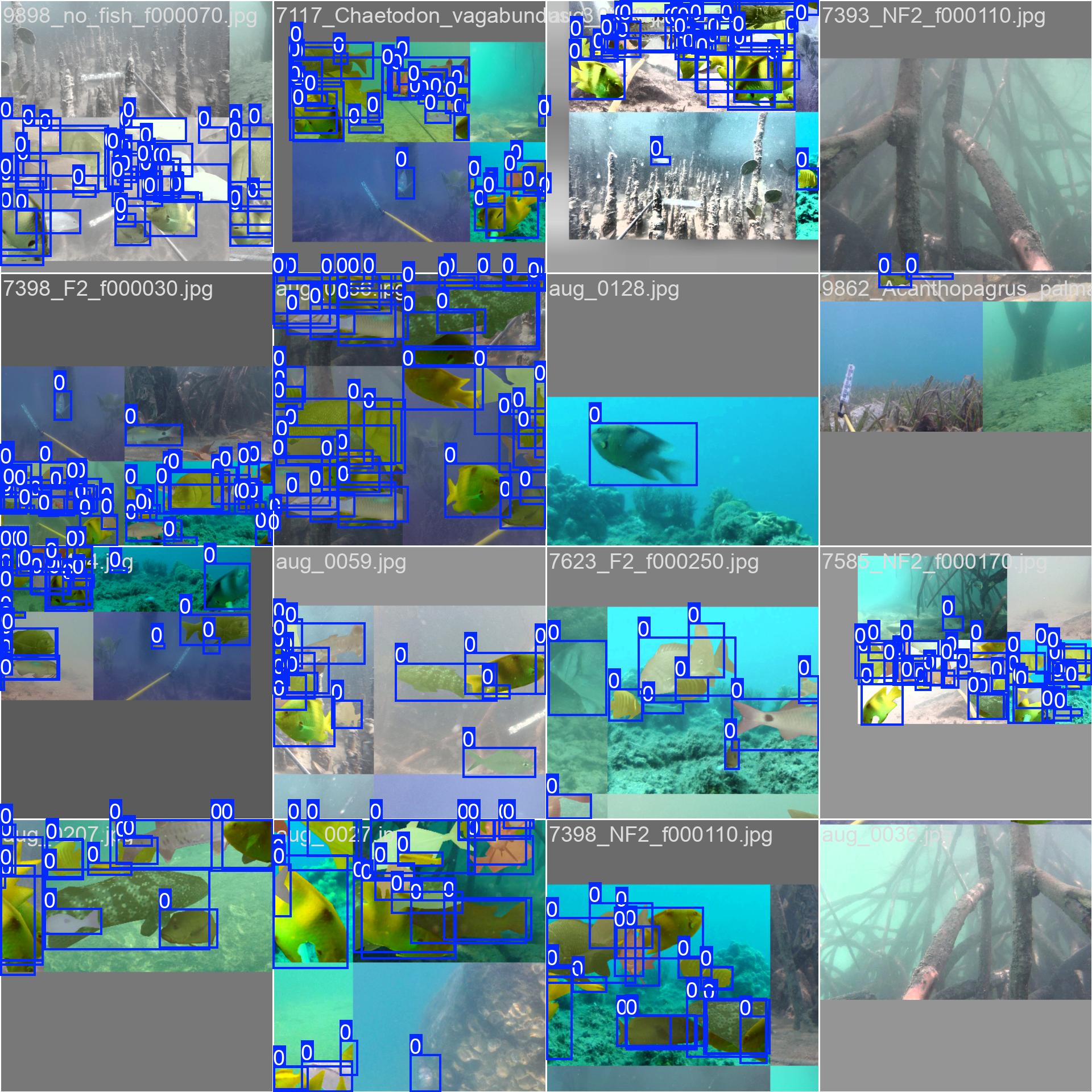} % Replace 'example-image' with your image file name
    \caption{Second training batch for PSADA model training run on DeepFish dataset.}
    \label{fig:your-label}
\end{figure}

\subsection{Live-Stream Test Data}
We collected 50 test images from a live-stream of natural fish habitats from a dock in Key West [source to link of youtube livestream], annotating these manually for object detection evaluation through RoboFlow. Notably, this dataset includes dense fish populations, which are underrepresented in DeepFish's training data.

%-------------------------------------------------------------------------
\section{Proposed Method}

For our baseline model we trained for 50 epochs, and each image was of size 640x640. For our second improved PSADA model, we trained for 28 epochs due to compute constraints and each image was of size 640x640.

\subsection{YOLO-Based Fish Detection}
We trained a YOLOv10 object detection model 
\cite{yolo} using the DeepFish bounding box annotations that were derived from provided DeepFish segmentation masks (see Figure 1). We used these provided black and white segmentation masks to estimate bounding boxes in green.

\subsection{Psuedo-Simulated Annealing Data Augmentation Algorithm}
Inspired by Deng's\cite{deng} image detection with humans using COCO [reference to what coco is], we implemented a modified optimization algorithm to improve predictions of bounding boxes in crowded scenarios that took up a large portion of our test data. Our novel algorithm is as follows, with our changes from the algorithm from Deng in \textbf{bolded text.}\\
\\
\begin{enumerate}
    \item \textbf{Select group centers.}
    \begin{itemize}
        \item Define a set \( C \) of group centers \( C = \{(x_1, y_1, s_1), \dots, (x_{|C|}, y_{|C|}, s_{|C|})\} \), where \( (x_i, y_i) \) are the coordinates and \( s_i \) is the normalized object size.
        \item The group number \( |C| \) is \textbf{sampled from a Poisson distribution} with parameter \( \lambda = 3 \), instead of a uniform range \([0, N]\), where \( N \) is a predefined hyperparameter.
        \item Group centers are selected by \textbf{random sampling of coordinates} within the base image dimensions, not directly from existing objects.
    \end{itemize}

    \item \textbf{Generate groups around each center.}
    \begin{itemize}
        \item For each group center \( c_i \in C \), generate a group \( \hat{G}_i \) of objects:
        \[
        \hat{G}_i = \{(x_1, y_1, s_1), \dots, (x_{|\hat{G}_i|}, y_{|\hat{G}_i|}, s_{|\hat{G}_i|})\}.
        \]
        \item The number of objects \( |\hat{G}_i| \) in each group is randomly sampled from the range \([1, M']\), where \( M' \) is proportional to \( M / |C| \), ensuring a balanced object distribution across groups.
        \item Enforce overlapping between each object \( g_{ij} \in \hat{G}_i \) and its corresponding group center \( c_i \) \textbf{using a simulated annealing approach} for optimal placement.
    \end{itemize}

    \item \textbf{Simulate realistic crowdedness.}
    \begin{itemize}
        \item Object sizes in a group follow a Gaussian distribution:
        \begin{equation}
        \begin{aligned}
        p(s_j \mid s_i, I)
          &= \frac{1}{\sqrt{2\pi\sigma^2}}\,
             \exp\left(-\frac{(s_j - s_i)^2}{2\sigma^2}\right).
        \end{aligned}
        \end{equation}
        where \( \sigma \) is a fixed standard deviation. \textbf{In the adjusted implementation, \( \sigma = 30 \) (pixel units)} instead of \( \sigma = 0.2 \) as a normalized value.

        \item Coordinates \( x \) and \( y \) are \textbf{refined iteratively using simulated annealing}, starting from uniform sampling around the group center \( (x_i, y_i) \) with displacements:
        \begin{equation}
        \begin{aligned}
        x_j &\sim U(x_i - \tau d_w,\; x_i + \tau d_w),\\
        y_j &\sim U(y_i - \varepsilon d_h,\; y_i + \varepsilon d_h).
        \end{aligned}
        \end{equation}
        where \( d_w \) and \( d_h \) are maximum allowable displacements. The degree of crowdedness is controlled via hyperparameters \( \tau \) and \( \varepsilon \), and placement temperature \( T \) decays geometrically over iterations:
        \begin{equation}
        \begin{aligned}
        T &\leftarrow T \cdot \gamma,\\
        \gamma &= 0.95.
        \end{aligned}
        \end{equation}
    \end{itemize}
\end{enumerate}

\subsubsection{Justification of Algorithm Adjustments}

Initially, we used a Poisson distribution to pick the number of center points (and thus number of groups) that are in the model. It was not feasible for us to pick the number of groups based on the number of objects in the image, because a large number of the training images had no fish. Additionally, the mode number of fish in a training image was one. Our method differs from the work of Deng et al. specifically because of this discrepancy -- we don't only lack clustering, but we lack multiple-object images in our training and validation data. Thus, if we wanted to introduce our neural network to clustering in the training set, particularly multiple clusters per image, we needed to pick a distribution. There is not much literature on the statistical modeling of animal groupings, so we made the choice to model the number of group centers with a Poisson distribution, which is often used for time series data. Each image is a snapshot into the natural environment of fish, so we can interpret the probabilistic model as the number of fish schools swimming in the area of the camera at some given time-interval. 
\\

Because we are not choosing groups based off of original images, naturally we will random sample the group centers via a uniform distribution. 
\\

Moreover we modify the algorithm for picking neighbors of the center image with a simulated annealing-style algorithm whose acceptance function is a negative exponential function based on initial temperature $\gamma$. This choice allowed us to keep our CNN robust to images with both individual fish and fish groupings. Initially, the probability of the image being far away from its center point is high, and with every new fish pasted to the image, this probability decreases. Thus, we are left with varied images with both dense objects and lone objects, as shown in Figure 2. 

%-------------------------------------------------------------------------
\section{Experimental Results}
\subsection{Training and Evaluation}
We trained two models:
\begin{enumerate}
    \item Baseline YOLOv10 model on the original DeepFish dataset. (what we are calling our "base model")
    \item YOLOv10 model with pseudo-simulated annealing data augmented algorithm and training on DeepFish and 500 additional images. (what we are calling our "PSADA model"). This is described by the algorithm in section 4.2.
\end{enumerate}

Training results are summarized in Figure 4(a) and (b). Observe for both models that training box loss decreased significantly from roughly 3.0 to 1.5 for both models, but PSADA reached this box loss minimum in less training time. We see the same trends for classification loss and distribution focal loss. Precision and recall increase for both, and we see analogous trends for similar decreasing losses for the validation as well. Notably, mAP50 and mAP50-95 reached up to 0.8 and 0.7 respectively for PSADA at a much faster rate than the baseline model. Other notable observations of training metrics include the validation boss loss oscillations for PSADA compared with more consistent declines in box loses for baseline, but this does not appear to affect the model performance later on. Furthermore, in the baseline model, all loss metrics increase for the first 10 epochs before a loss decrease begins, but this does not occur for PSADA.

\begin{figure}[htbp]
    \centering
    % First subplot
    \begin{subfigure}[b]{1\linewidth} % Adjust width as needed
        \centering
        \includegraphics[width=\linewidth]{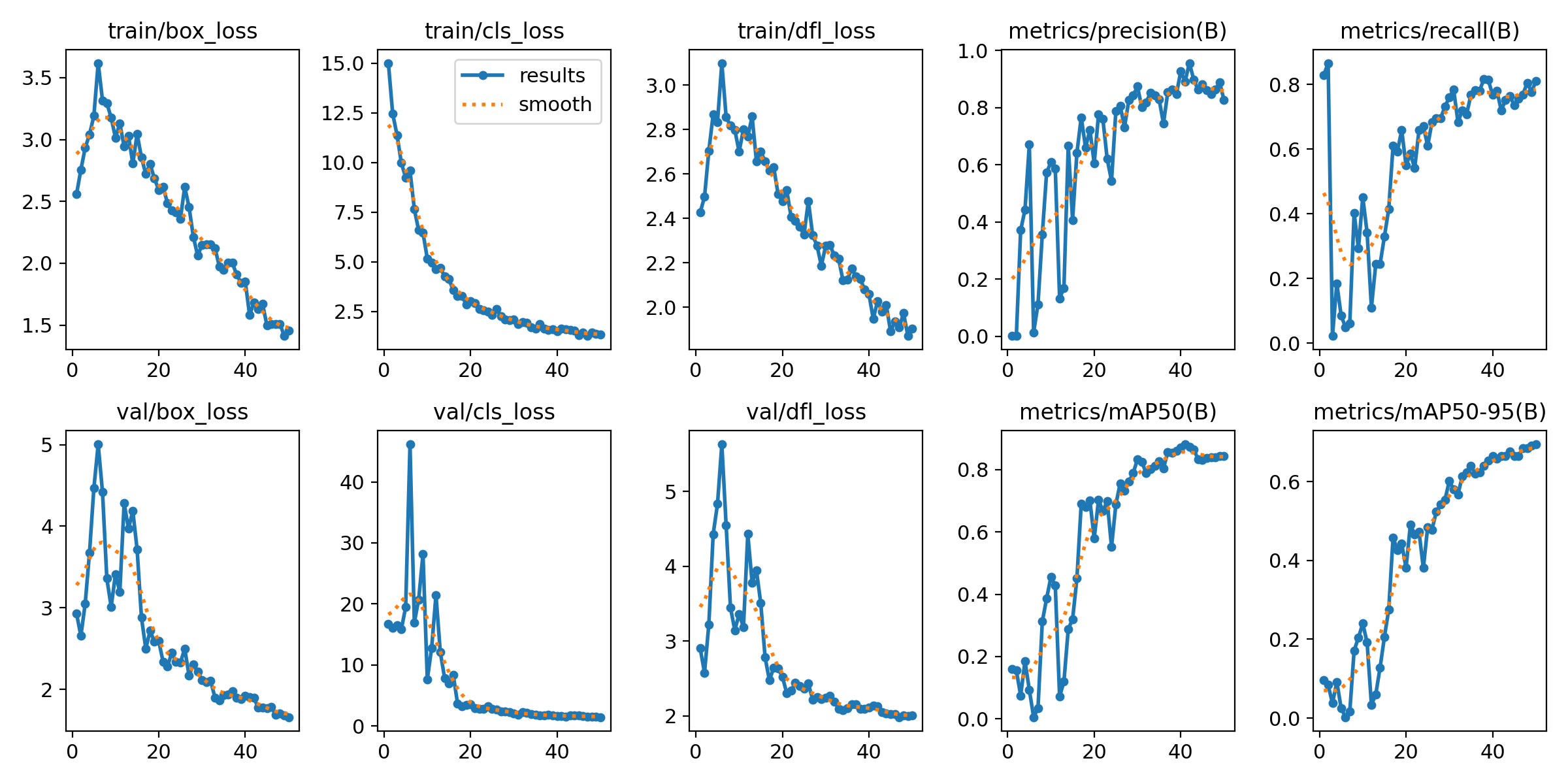} % Replace with your image
        \caption{Base model training results for 50 epochs.}
        \label{fig:top-subplot}
    \end{subfigure}
    
    % Vertical spacing between subplots
    \vspace{1em}

    % Second subplot
    \begin{subfigure}[b]{1\linewidth} % Adjust width as needed
        \centering
        \includegraphics[width=\linewidth]{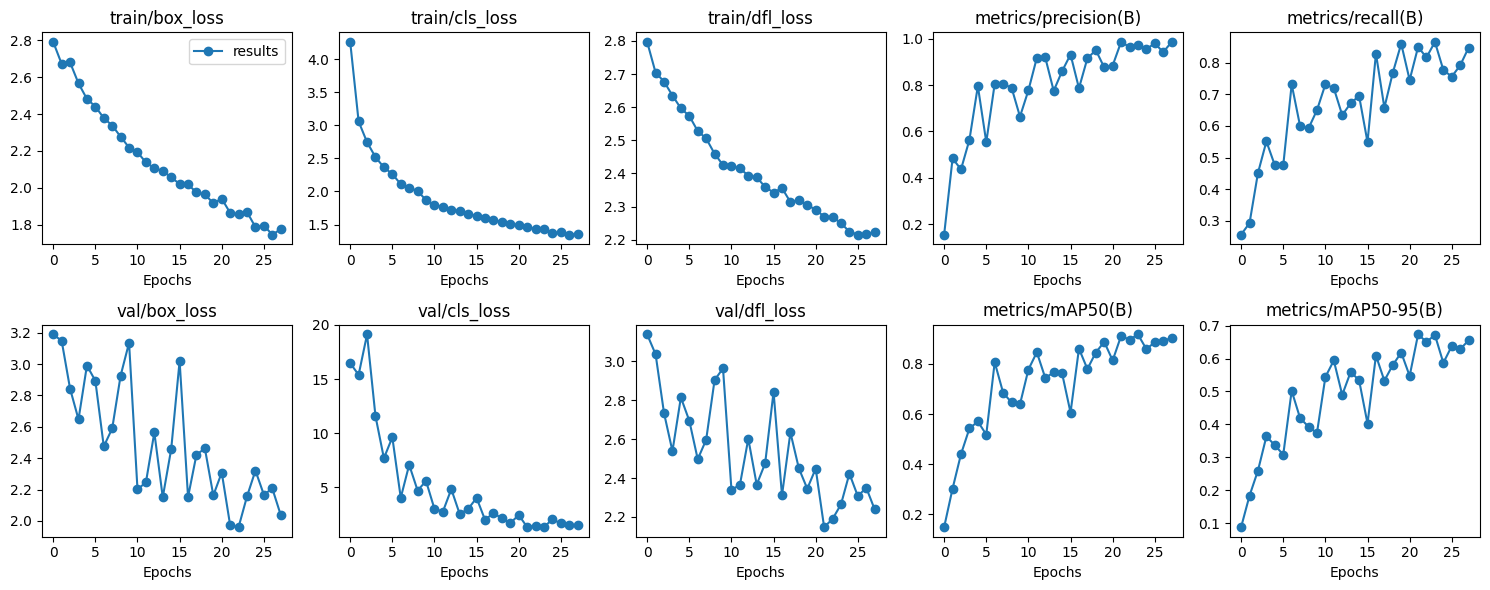} % Replace with your image
        \caption{PSADA model training results for 28 epochs.}
        \label{fig:bottom-subplot}
    \end{subfigure}

    % Main figure caption
    \caption{Training results for both models. From left to right, starting at the top row: (1) training box loss, (2) training classification loss, (3) training distribution focal loss (DFL), (4) precision (metrics/precision), and (5) recall (metrics/recall). The bottom row shows: (6) validation box loss, (7) validation classification loss, (8) validation DFL loss, (9) mean Average Precision at IoU threshold 0.5 (mAP50), and (10) mean Average Precision over IoU thresholds from 0.5 to 0.95 (mAP50-95).}
    \label{fig:overall-figure}
\end{figure}

\subsection{Results}

The final results of this paper are the testing our Baseline YOLO model and our PSADA YOLO model on the detection in our novel Florida fish dataset from Florida Keys livestream data. Upon running our models on set of Florida fish images, a set of bounding boxes were outputted that estimated the expected location of a fish. We counted the number of such bounding boxes relative to the ground truth bounding boxes from our manual annotations and ultimately calculated the number of detected fish by both models compared to ground truth. These results are depicted in Figure 7. On average, PSADA was able to detect more than double the fish in the Florida dataset than our baseline YOLO model. Compared with the ground truth fish count, on average the baseline model detected less than a quarter of fish, whereas the PSADA detected nearly half of the ground truth count. The difference in this fish count via bounding boxes is displayed in Figure 5, in which PSADA is able to detect a high proportion of crowded fish compared with poor detection of any fish in crowded schools by the baseline model. Qualitatively we observed that both models performed well when fish were isolated in uncrowded environments. Furthermore, to quantitatively compare the performance of our two models on the Florida keys dataset, we calculated Intersection over Union (IoU) scores to determine how well each model accurately detected the boxed location of the fish on average. We display the IoU distribution in Figure 6 comparing our baseline and PSADA models. This score quantifies the overlap between the predicted bounding box with the ground truth bounding boxes, and we observe that the overlap distribution is more skewed left for the PSADA model with a larger proportion of higher IoU scores than baseline.

%-------------------------------------------------------------------------
\section{Conclusion}
We presented a novel YOLO-based fish detection model tailored for real-world natural habitats, addressing challenges of variable lighting, water conditions, and crucially, dense fish populations. By generating bounding boxes from DeepFish segmentation masks and integrating a simulated annealing-based optimization algorithm, we achieved significant improvements in crowded object detection from our baseline. Our results demonstrate the efficacy of our method, particularly on live-streamed fish habitat data. This project was limited by the training constraints of Google Colab and by constraints in access to GPUs and compute. We believe that future work in using more GPU-intensive training for the PSADA algorithm could yield improved results. Furthermore, even though our PSADA model was significantly better than the baseline YOLO model at detecting fish in the Florida keys dataset, it still missed almost half of the fish present from our manual annotations as demonstrated by Figure 7. There are various possibilities for this aside from training time. 
\begin{figure}[htbp]
    \centering
    % First subplot
    \begin{subfigure}[b]{0.8\linewidth} % Adjust width as needed
        \centering
        \includegraphics[width=\linewidth]{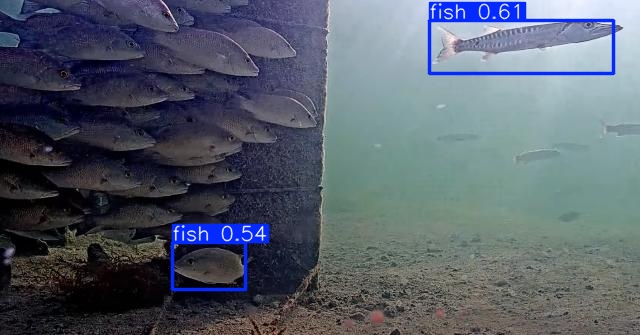} % Replace with your image
        \caption{Base model performance on Florida Keys image of crowded fish from formal model testing run.}
        \label{fig:top-subplot}
    \end{subfigure}
    
    % Vertical spacing between subplots
    \vspace{1em}

    % Second subplot
    \begin{subfigure}[b]{0.8\linewidth} % Adjust width as needed
        \centering
        \includegraphics[width=\linewidth]{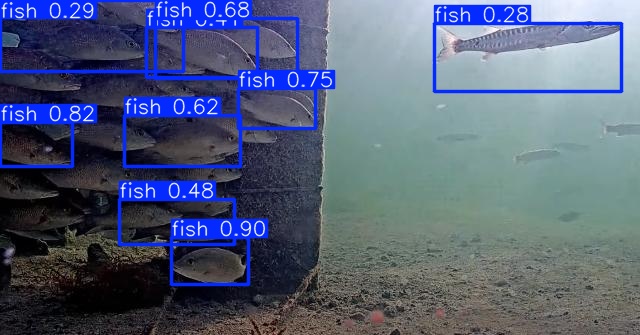} % Replace with your image
        \caption{PSADA model performance on Florida Keys image of crowded fish from formal model testing run.}
        \label{fig:bottom-subplot}
    \end{subfigure}

    % Main figure caption
    \caption{Our annealing-based data augmentation algorithm model, the PSADA model, (b) outperforms our baseline model (a) in detecting crowded fish and fish in unpredictable natural environments.}
    \label{fig:overall-figure}
\end{figure}
First, the DeepFish dataset on which we trained is a dataset of Australian fish, which are not the same species in our Florida keys dataset. For example, in the DeepFish dataset the fish tend to be similar in size and features, whereas the Florida keys dataset that we generated included sharks and barracuda fish that were sometimes missed by our model. Second, DeepFish dataset rarely included images of crowded fish and typically most images displayed only one or two fish with little to no crowding from our qualitative assessment, so training on this dataset and subsequently testing on a highly crowded fish dataset such as the Florida keys made it difficult for PSADA to detect all fish. As can be seen in Figure 5, though PSADA significantly improved in detection of crowded fish from our baseline, but due to the high volume of fish in certain crowded environments, PSADA was unable to detect fish in certain areas of maximal crowding such as the bottom left of Figure 5(b). Future work in fine tuning our model and further testing of possible augmentations to our algorithm is another next step. Lastly, adding a classification element to the training of PSADA for the identification of different fish species in crowded environments is an area of future research that could be immensely beneficial for ecological researchers studying fish counts and marine and coral health.

\section{Author Notes}
We as authors of this paper affirm here that we split the research and writing for this report evenly and collaborated successfully throughout the duration of the report. Additionally, we are considering publishing this work upon further model training and would appreciate feedback on how to improve our paper for submission to a journal. We further would like to acknowledge and thank Professor Zickler for his inspiring instruction on computer vision and feedback on this project.

We sourced our live Florida keys fish data from images taken of the Viva The Keys publically available underwater live camera found at this link  
\href{https://www.youtube.com/watch?v=qi0mY6zVQnY}{here} which regularly changes in setting, angle, lighting and displayed fish populations in the Florida Keys.

\begin{figure}[htbp]
    \centering
    \includegraphics[width=\linewidth]{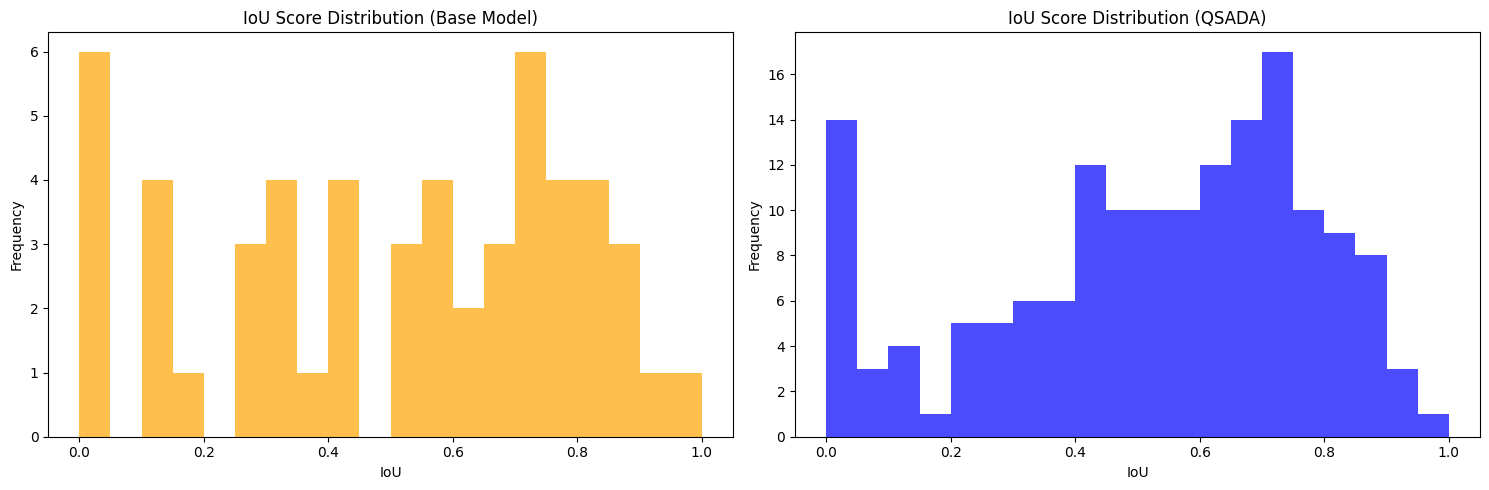} % Replace 'example-image' with your image file name
    \caption{IoU distribution for Base model and PSADA model on Florida Keys Dataset predicted fish bounding boxes vs. ground truth bounding boxes.}
    \label{fig:your-label}
\end{figure}

\begin{figure}[htbp]
    \centering
    \includegraphics[width = \linewidth]{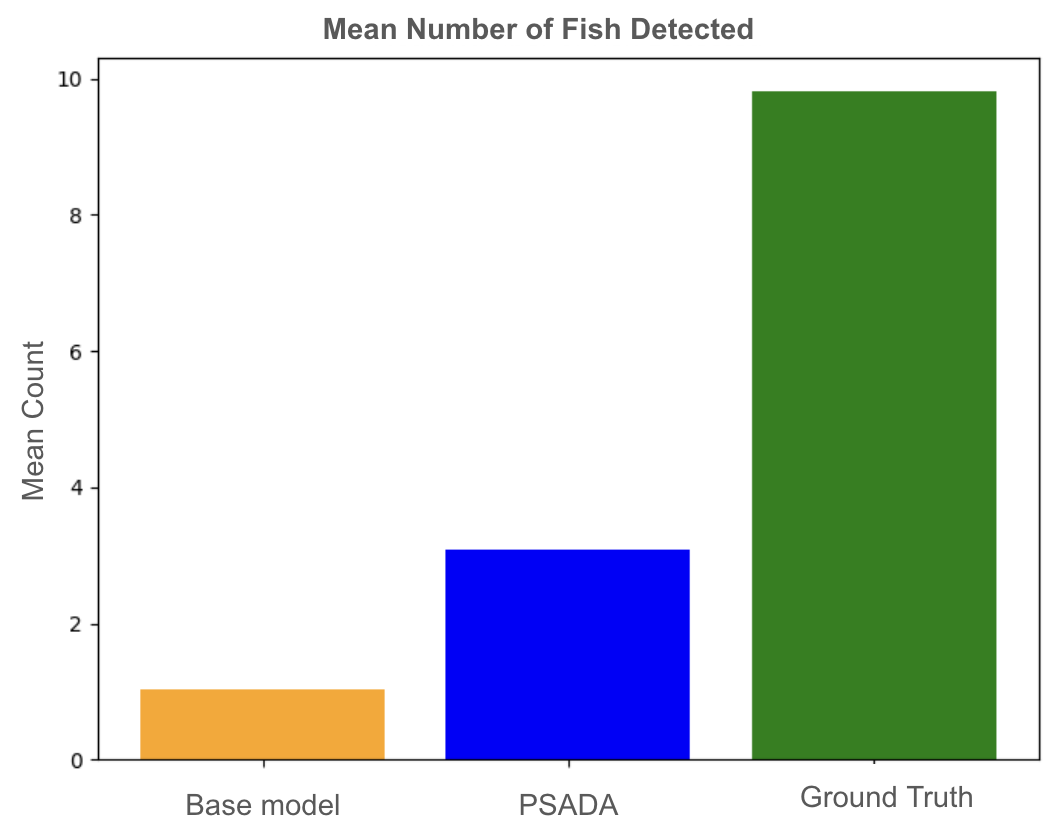} % Replace 'example-image' with your image file name
    \caption{Mean number of fish detected in Florida Keys Dataset by each model vs. ground truth. }
    \label{fig:your-label}
\end{figure}

{\small
\bibliographystyle{ieee}
\bibliography{egbib}
}

\end{document}